\documentclass[journal]{IEEEtran}

\usepackage{xcolor,soul,framed} %,caption

\colorlet{shadecolor}{yellow}
\usepackage[pdftex]{graphicx}
\graphicspath{{../pdf/}{../jpeg/}}
\DeclareGraphicsExtensions{.pdf,.jpeg,.png}

\usepackage[cmex10]{amsmath}
\usepackage{amssymb}
\usepackage{array}
\usepackage{mdwmath}
\usepackage{mdwtab}
\usepackage{eqparbox}
\usepackage{url}
\usepackage{amsmath}
\usepackage{caption,subcaption}

\usepackage{hyperref}

\usepackage{algorithm} 
\usepackage{algpseudocode} 
\usepackage{array}%需要用到该宏包
\usepackage{footnote}
\makesavenoteenv{tabular}
\usepackage{float}
\usepackage{subfloat}
\begin{document}
% \bstctlcite{IEEEexample:BSTcontrol}
    \title{Pre-trained Transformer-Enabled Strategies with Human-Guided Fine-Tuning for End-to-end Navigation of Autonomous Vehicles}
  \author{Dong Hu,   Chao Huang,~\IEEEmembership{Senior Member,~IEEE,}  Jingda Wu, \\ ~and Hongbo Gao,~\IEEEmembership{Member,~IEEE}

  \thanks{Corresponding author: Chao Huang.}
  \thanks{Dong Hu, Chao Huang and Jingda Wu are with the Department of Industrial and Systems Engineering, the Hong Kong Polytechnic University, Hong Kong (E-mail: dong24.hu@connect.polyu.hk;
hchao.huang@polyu.edu.hk; jingda.wu@polyu.edu.hk).}

\thanks{Hongbo Gao is with Department of Automation, School of Information Science and Technology, and Institute of Advanced Technology, University of Science and Technology of China, Hefei 230026, China, and also with Nanyang Technological University, 639798, Singapore (e-mail: ghb48@ustc.edu.cn and eee-hbgao@ntu.edu.sg).}

}

% The paper headers
\markboth{}{xx \MakeLowercase{\textit{et al.}}: xx}

% ====================================================================
\maketitle

% === ABSTRACT ====================================================================

\begin{abstract}
%\boldmath

Autonomous driving (AD) technology, leveraging artificial intelligence, strives for vehicle automation. End-to-end strategies, emerging to simplify traditional driving systems by integrating perception, decision-making, and control, offer new avenues for advanced driving functionalities. Despite their potential, current challenges include data efficiency, training complexities, and poor generalization. This study addresses these issues with a novel end-to-end AD training model, enhancing system adaptability and intelligence. The model incorporates a Transformer module into the policy network, undergoing initial behavior cloning (BC) pre-training for update gradients. Subsequently, fine-tuning through reinforcement learning with human guidance (RLHG) adapts the model to specific driving environments, aiming to surpass the performance limits of imitation learning (IL). The fine-tuning process involves human interactions, guiding the model to acquire more efficient and safer driving behaviors through supervision, intervention, demonstration, and reward feedback. Simulation results demonstrate that this framework accelerates learning, achieving precise control and significantly enhancing safety and reliability. Compared to other advanced baseline methods, the proposed approach excels in challenging AD tasks. The introduction of the Transformer module and human-guided fine-tuning provides valuable insights and methods for research and applications in the AD field.

\end{abstract}

% === KEYWORDS ====================================================================
% =================================================================================

\IEEEpeerreviewmaketitle

% === I. INTRODUCTION =================================
\textbf{\section{Introduction}}

\IEEEPARstart{T}{he} rapid technological progress has elevated autonomous driving (AD) to a pivotal focus in the automotive industry. Traditional AD systems, categorized into three stages—perception, decision-making, and execution—rely on intricate modules and complex rules \cite{chib2023recent}. Yet, these methods often grapple with achieving optimal performance amid complex traffic environments and uncertainties. In recent years, end-to-end learning emerges as a disruptive approach, breaking from traditional modular design by integrating perception, decision-making, and control into a unified system \cite{feng2023dense}. 

There are two main types of end-to-end solutions: imitation learning (IL) and reinforcement learning (RL). IL assumes that expert trajectories represent optimal behavior, learning an approximate expert driving policy by replicating expert actions in given states. Early IL methods like conditional imitation learning (CIL) \cite{codevilla2018end} and CILRS \cite{codevilla2019exploring} employed a conditional architecture. However, IL faces limitations in supervised learning (SL) capabilities and distribution shift problems, relying solely on static expert demonstration datasets. In contrast, RL learns from the agent's interaction with its environment, handling large state, action spaces and complex scenarios. Despite challenges such as low data efficiency and safety concerns, RL benefits from human knowledge in feature engineering. This aids in selecting task-relevant features, simplifying the state space, reducing problem complexity, and allowing the model to focus on critical task information. An effective architecture involves continuous learning using RL after pre-training. In \cite{zou2021deep}, a behavior cloning (BC) method based on expert experience pre-trains deep deterministic policy gradient (DDPG) agents, initiating RL end-to-end control policies near optimal strategies to hasten convergence. Additionally, the DAgger algorithm is considered for improved pre-training policies \cite{shi2023efficient}. Despite enhanced training efficiency, these approaches grapple with safety and generalization challenges.

Currently, significant research focuses on large language models (LLMs) like ChatGPT \cite{wu2023brief}, which exhibit powerful language and context understanding capabilities. These capabilities show potential for integration with decision-making systems in AV \cite{wang2023bevgpt, xu2023drivegpt4}. ChatGPT, utilizing a pre-trained model based on the Transformer architecture, achieves outstanding generation capabilities through fine-tuning by RL from human feedback (RLHF). Drawing inspiration from ChatGPT's training process, we propose a novel end-to-end strategy training architecture for AD. First, we employ a actor network based on the Transformer module for BC pre-training. Then, this network undergoes further fine-tuning using RL with human guidance (RLHG) to adapt to specific driving environments and tasks. The fine-tuning process involves interaction with humans, guiding the model to learn precise and safe driving decisions through their guidance and feedback. We named this strategy framework PTA-RLHG. This combination of pre-training and fine-tuning enables the end-to-end policy to glean more intelligent strategies from extensive semantic image data, significantly enhancing driving strategy performance. The contributions of this paper can be summarized as follows:

\begin{enumerate}
    \item Integrating the Transformer architecture into the actor network of RL, coupled with pre-training through BC, enhances the model's representation ability by advancing context understanding for driving tasks.
    \item Fine-tuning with RLHG, following the pre-trained Transformer-enabled actor (PTA) network, allows the model to systematically optimize its driving strategy through human supervision, intervention, demonstration, and reward feedback.
    \item In simulated highway and urban scenarios, we compare our proposed approach with state-of-the-art RL and IL baseline methods. Our method clearly outperforms in addressing challenging AD tasks comprehensively.
\end{enumerate}

This end-to-end control strategy provides new ideas and methods for the future development of intelligent transportation systems. The remainder is organized as follows: in Section II, we discuss relevant literature, followed by a detailed explanation of our proposed methodology in Section III. The experimental details is outlined in Section IV, with results and analysis presented in Section V. Lastly, major conclusions are provided in Section VI.

% === II. Methodology Overview========================
\textbf{\section{Related Work}}

\subsection{Transformer-based Driving Strategy}

The attention mechanism, initially prominent in natural language processing \cite{vaswani2017attention}, has evolved into a powerful tool across various deep learning domains. Its success in computer vision, notably with the vision Transformer (VIT) \cite{dosovitskiy2020image, qian2021blending}, showcases its adaptability. In ImageNet classification, attention aids models in autonomously learning crucial image regions, leading to outstanding performance. This influence extends to AD research, covering motion prediction \cite{gao2020vectornet,huang2023differentiable,huang2023conditional}, driver attention prediction \cite{gou2022driver}, and object tracking \cite{meinhardt2022trackformer}. In end-to-end AD, Transferuser \cite{chitta2022transfuser} employs a multimodal transformer for integrating forward camera images and LiDAR point clouds. NEAT \cite{chitta2021neat} uses an intermediate attention map to iteratively compress 2D image features into a Bird's Eye View (BEV) representation. Interuser \cite{shao2023safety} leverages a Transformer for fusing and processing multimodal, multi-view sensors, achieving a comprehensive scene understanding. ReasonNet \cite{shao2023reasonnet} utilizes a Transformer for inferring temporal behaviors, effectively handling interactions and relationships between features in different frames, aiding in detecting adverse events, particularly predicting potential dangers from obstructed objects.

The Transformer, serving as the core structure with a self-attention mechanism in LLMs, undergoes optimization and training for application in AD tasks. BEVGPT \cite{wang2023bevgpt}, using BEV images as the sole input, employs a generative pre-trained large model for driving decisions based on traffic scenes. DriveGPT4 \cite{xu2023drivegpt4} adopts an interpretable end-to-end AD system, providing explanations for vehicle behavior and corresponding reasoning. DiLU \cite{wen2023dilu}, with reasoning and reflective modules, makes decisions based on common knowledge and demonstrates superior generalization over RL-based approaches. Other research directions focus on multimodal large model simulation or world model construction. DriverDreamer \cite{wang2023drivedreamer}, the first real-world driving scenario world model, uses a powerful diffusion model to represent complex environments comprehensively, predicting future states in subsequent stages. Despite challenges in computational resources, safety, and interpretability in applying LLMs to AD, the integration of state-of-the-art AI technology with AD holds significant potential for the future of intelligent transportation.

\begin{figure*}
    \begin{center}
    \includegraphics[width=0.95\linewidth]{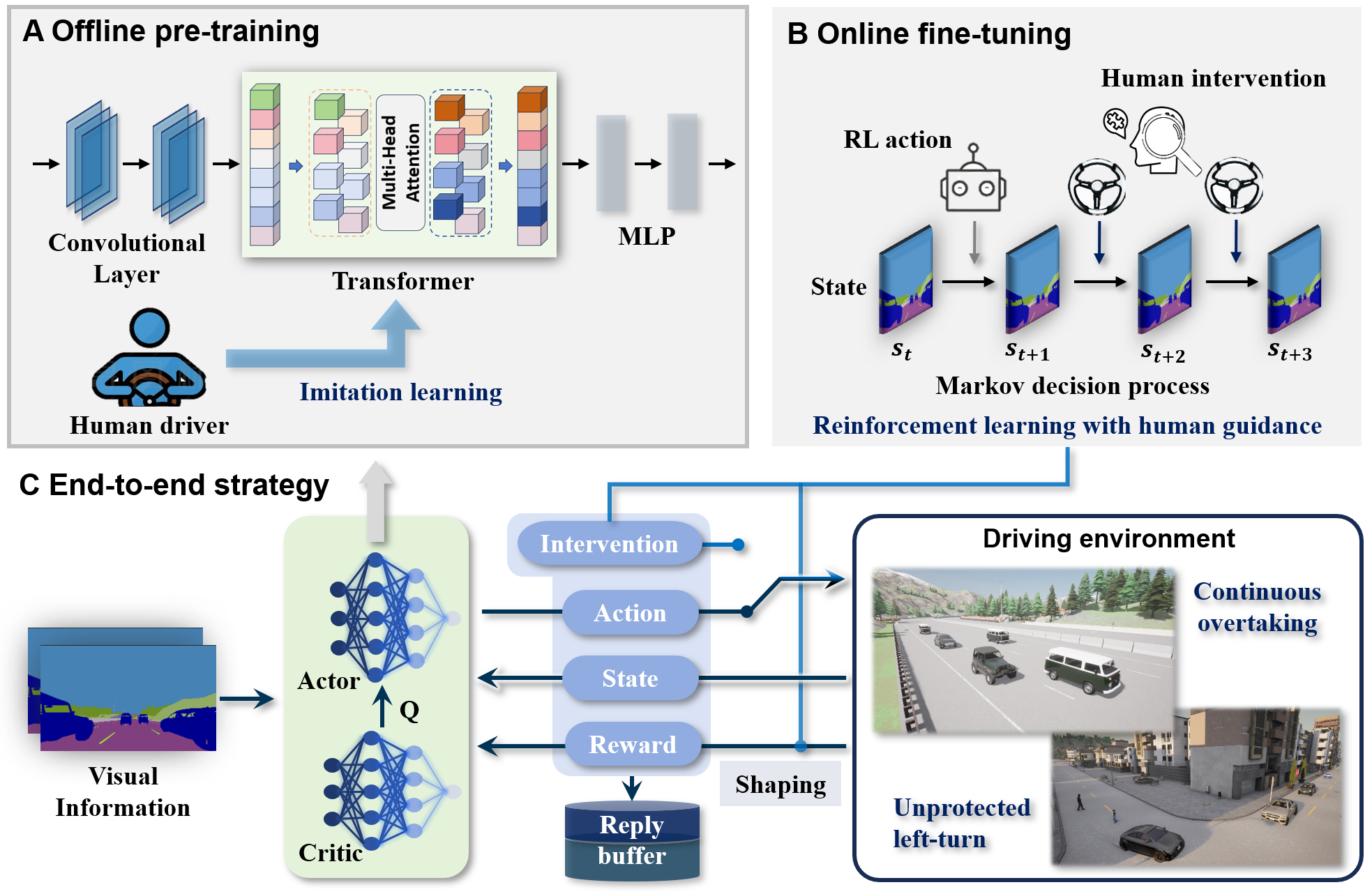}
    \caption{The architecture of the proposed end-to-end strategy: (A) represents the PTA network is pre-trained through imitation learning; (B) represents RLHG for online fine-tuning; and (C) represents the interaction process between RL and the environment.} \label{Framework}
 \end{center}                                       
\end{figure*}

\subsection{Reinforcement Learning with Human Knowledge}

In recent years, significant progress has been made in RL within machine learning. RL involves agents interacting with an environment, utilizing the Bellman equation to maximize cumulative reward. However, slow data collection, particularly inefficient updates of extensive policies and evaluation networks, constrains deep reinforcement learning (DRL). Common exploration techniques like $\epsilon$-greedy \cite{mnih2015human} and noise injection \cite{lilicrap2016continuous} are used during training to approach the global optimum policy. Nevertheless, these methods may lead to unsafe behaviors, causing significant cost losses. Hence, RL faces challenges in efficiency and safety. Transfer RL \cite{HU2024130097} and curriculum RL \cite{hu2023asynchronous} aim to accelerate learning by leveraging prior knowledge effectively. Model-based RL \cite{wu2022uncertainty} benefits from planning in simulated environments, reducing trial-and-error.

Leveraging human prior knowledge significantly enhances RL by improving training efficiency and safety, effectively reducing model uncertainty in the early stages \cite{HU2023121227}. Strategies integrating expert experiences directly into RL training data, as seen in scenarios like city roundabouts \cite{liu2022improved}, expedite training. An approach proposed in \cite{pfeiffer2018reinforced} introduces an IL pre-training constrained policy optimization algorithm, followed by continuous RL training. Combining BC and generative adversarial IL (GAIL), \cite{tai2018socially} enhances socially-driven path planning performance. In \cite{huang2023goal}, an expert model pre-trained using IL on scene encoding assists agents. Human prior knowledge guides RL behavior through well-designed reward shaping, directing the model to learn in the desired direction and avoiding unsafe or task-inconsistent behavior. Introducing a curiosity-driven reward shaping, \cite{wu2021deep} improves the data efficiency of RL in lane keeping and overtaking tasks.

Researchers actively pursue effective regulations and oversight measures for ensuring the safe and reliable operation of RL systems in practical applications. \cite{WU202375} proposes a DRL approach employing real-time human guidance to enhance efficiency and performance by seamlessly transferring control between humans and agents. \cite{9793564} introduces a unique priority experience replay strategy and human-intervention-based reward shaping, further boosting efficiency. RLHF, as an interactive method, directly utilizes human feedback as a strategy label to maximize information acquisition efficiency \cite{griffith2013policy, knox2011augmenting}. Overall, in terms of data efficiency, human intervention and feedback are anticipated to surpass typical knowledge injection RL, though challenges like human fatigue necessitate further optimization.

% === 3. method =================================================================================
\textbf{\section {Methodology}}
% ---------------------------------------------------------------
\subsection{Framework}

This study proposes a novel AV end-to-end navigation framework called PTA-RLHG, as illustrated in Fig. \ref{Framework}, consisting of three main components. Firstly, the Transformer module is introduced into the actor network, providing global attention to the input image representation sequence. This fosters the establishment of contextual relationships and the capture of dependencies at different scales within the sequence. To comprehensively train the Transformer module, we employ the BC algorithm with human demonstration data for pre-training. This allows the actor model to learn universal representations and obtain reasonable initial parameter settings, facilitating faster convergence and adaptation to task-specific training data.

Following pre-training, online fine-tuning of the policy network occurs through RLHG. This process involves human supervision, intervention, demonstrations, and feedback on the value of agent actions through reward shaping. Such a design accelerates the fine-tuning process, preventing the model from getting stuck in local optima. By leveraging RL's exploration mechanism and adhering to the principle of maximizing cumulative rewards, the agent's output actions can surpass human demonstration levels. This integrated approach leads to significant performance improvements in the AV field.

% ---------------------------------------------------------------
\subsection{Pre-trained Transformer-enabled Actor}

The actor network comprises a 2-layer convolutional encoder, Transformer module, and multi-layer perception (MLP) decoder, forming the structural foundation. Illustrated in Fig. \ref{Framework} (A), the input image data undergoes initial processing through the first-layer convolutional network. In this stage, convolutional operations are utilized to extract features from raw data, enabling the capturing of local patterns and spatial relationships inherent in the input data. Subsequently, the output undergoes further refinement in the second-layer convolutional network. By stacking multiple layers of convolutional layers and applying non-linear activation functions, this stage learns abstract features, gradually reducing data size. Following the encoding stages, the output is directed to the Transformer module. Comprising self-attention mechanisms and feedforward neural networks, this component can globally attend to and model input sequences. Through self-attention mechanisms, the model establishes global contextual relationships within the feature representations from the encoder, effectively capturing long-range dependencies within the sequence. Finally, the processed output from the Transformer module is conveyed to the MLP decoder, responsible for transforming high-level feature representations into final predictions or generating outputs. This integrated structure optimally combines the strengths of convolutional and self-attention mechanisms, enabling efficient feature learning and the capture of complex patterns at various levels of input data.

The self-attention mechanism from Transformer is employed here to incorporate global context into the input image, leveraging their complementary nature. The input sequence in the Transformer architecture consists of discrete tokens, with each token is characterized by a feature vector. Spatial inductive bias is incorporated by applying positional encoding to the feature vectors. Mathematically, we denote the input sequence as $\boldsymbol{F}^{in} \in \mathbb{R}^{M \times D_f}$, where $M$ represents the number of tokens in the sequence, and each token is described by a feature vector with a dimension of $D_f$. The Transformer calculates sets of queries, keys, and values through linear projections \cite{9578103}:

\begin{equation}
\boldsymbol{Q}=\boldsymbol{F}^{i n} \boldsymbol{W}^q, \boldsymbol{K}=\boldsymbol{F}^{i n} \boldsymbol{W}^k, \boldsymbol{V}=\boldsymbol{F}^{i n} \boldsymbol{W}^v
\end{equation}

\noindent where $\boldsymbol{W}^q$, $\boldsymbol{W}^k$ and $\boldsymbol{W}^v$ are weight matrix.

The multi-head attention mechanism serves as the core of the proposed PTA structure, extracting information from various representation subspaces by concurrently computing multiple independent self-attention mechanisms. In the Transformer, the formula for the multi-head attention mechanism is as follows:

\begin{equation}
\operatorname{MultiHead}(\boldsymbol{Q}, \boldsymbol{K}, \boldsymbol{V})=\operatorname{concat}\left(\text{head}_1, \cdots, \text{head}_h\right) \boldsymbol{W^O}
\end{equation}

\noindent where the attention results from $h$ heads are concatenated to capture different aspects of information, and $h$ is set to 4 in this study. $\text{head}_i=\operatorname{Attn}(\boldsymbol{Q_i}, \boldsymbol{K_i}, \boldsymbol{V_i})$ represents the calculation result of the i-th attention head. $\boldsymbol{W^O}$ is the output weight matrix for the multi-head attention.

The self-attention mechanism calculates dependencies between different positions in the input sequence and assigns weights to each position. In the Transformer encoder, attention weights are determined by the scaled dot-product between $\boldsymbol{Q}$ and $\boldsymbol{K}$, followed by the aggregation of values corresponding to each query \cite{vaswani2017attention}.

\begin{equation}
\operatorname{Attn}(\boldsymbol{Q}, \boldsymbol{K}, \boldsymbol{V})=\operatorname{softmax}\left(\frac{\boldsymbol{Q} \boldsymbol{K}^{\top}}{\sqrt{d_k}}\right) \boldsymbol{V}
\end{equation}

\noindent where $d_{k}$ is the dimensionality of the key.

Afterwards, Transformer employs a non-linear transformation $\operatorname{MLP}(\cdot)$ to compute the output features $\boldsymbol{F}^{o ut}$, which has the same shape as the input features $ \boldsymbol{F}^{i n}$.

\begin{equation}
\boldsymbol{F}^{o ut} = \operatorname{MLP}(\operatorname{Attn}(\boldsymbol{Q}, \boldsymbol{K}, \boldsymbol{V}))+\boldsymbol{F}^{i n}
\end{equation}

% BCCCCCCCCCC

Pre-training the actor network based on the Transformer can be achieved by imitating the behavior policy of actual human participants. Therefore, we utilize human demonstration data to perform BC for pre-training the actor model. It's important to note that the pre-trained model is not intended to precisely mimic expert-level human behavior. Its primary purpose is to enable the model to learn general representations and obtain relatively reasonable initial parameter settings, facilitating faster convergence and adaptation to specific task training data. In practice, human performance can fluctuate with changes in mental and physical states. Our goal is to provide the Transformer with sufficient update gradients, so expert-level labels are not necessary. Essentially, the pre-trained model for the actor aims to offer approximately correct initial representations for RL agents. Let {\text{PA}} denote the pre-trained actor, and the training objective is to minimize the difference between the policy $\pi^{\text{PA}}$ and the human policy $\pi^{\text{H}}$. Therefore, the loss function is:

\begin{equation}
\mathcal{L}^{\text{PA}}(\theta^\pi)=\mathbb{E}\left[\left\|a_t^{\text {H}}-\pi^{\text{PA}}\left(s_t \mid  \theta^\pi \right)\right\|^2\right]
\end{equation}

\noindent where $\mathbb{E}(\cdot)$ is expectant of function, $a_t^{\text {H}}$ is the action provided by humans. And update the model using the gradient method as follows:

\begin{equation}
\pi_{\text{new}}^{\text{PA}} \leftarrow \pi_{\text{old}}^{\text{PA}}-lr^{\pi^{\text{PA}}} \cdot \nabla_{\theta^\pi} \mathcal{L}^{\text{PA}}(\theta^\pi)
\end{equation}

\noindent where $\leftarrow$ represents update, $lr$ is the learning rate of BC, and $\nabla$ is gradient. Through the aforementioned updates, the model $\pi_{\text{new}}^{\text{PA}}$ can learn general representations and appropriate initial parameters.

% ---------------------------------------------------------------
\subsection{Twin Delayed Deep Deterministic Policy Gradient}

In conventional applications of DRL, such as AV, the agent's control can be modeled as a Markov decision process (MDP), denoted by a tuple $\mathcal{M}$, as outlined below:

\begin{equation}
\mathcal{M}=(\mathcal{S}, \mathcal{A}, \mathcal{T}, \mathcal{R})
\end{equation}

\noindent where $\mathcal{S} \in \mathbb{R}^n$ is the state space, $\mathcal{A} \in \mathbb{R}^m$ is the action space (with $n$ and $m$ being the dimensions; $\mathbb{R}$ denotes the set of real numbers). And transition model is $\mathcal{T}: \mathcal{S} \times \mathcal{A} \to \mathcal{S}$, reward function is $\mathcal{R}: \mathcal{S} \times \mathcal{A} \to \mathcal{R}$.

At time step $t$, the agent performs action $a_t$ within state $s_t$ and obtains a reward signal $r_t = \mathcal{R}(s_t, a_t)$. Following this, the environment undergoes a transition to the subsequent state $s_{t+1}$ according to the dynamic transition $\mathcal{T}(\cdot|s_t,a_t)$. In the AD context, crafting an accurate transition probability model $\mathcal{T}$ for environmental dynamics can pose challenges. Hence, we utilize model-free RL to tackle this problem, removing the necessity to model transition dynamics.

The TD3 algorithm belongs to the actor-critic (AC) framework and is a deterministic DRL algorithm that combines elements of DDPG and double Q-learning. It has demonstrated good performance across multiple continuous control tasks. The TD3 algorithm involves six networks: the actor network $\pi(\cdot| \theta^{\pi})$, critic 1 network $Q_1(\cdot| \theta^{Q_1})$, critic 2 network $Q_2(\cdot| \theta^{Q_2})$, actor target network $\pi^{\prime}(\cdot| \theta^{\pi^{\prime}})$, critic 1 target network $Q_1^{\prime}(\cdot| \theta^{Q_1^{\prime}})$, and critic 2 target network $Q_2^{\prime}(\cdot| \theta^{Q_2^{\prime}})$. The TD3 algorithm establishes two independent critic networks, and when computing the target values, it selects the minimum value between the two to mitigate the problem of overestimation in the networks.

\begin{equation}
R_t=r_t+\gamma \min _{i=1,2} Q_i^{\prime}\left(S_{t+1}, \tilde{A} \mid \theta_{Q_i^{\prime}}\right)
\end{equation}

\noindent where $r_t$ denotes the instantaneous reward at time step $t$, $\gamma$ is the discount factor, $R_t$ represents the total cumulative reward, $s_{t+1}$ is the state at time $t+1$, and $\tilde{A}$ denotes the target action with additional noise.

Deterministic policies face a challenge of overfitting to minimize peaks in value estimates. Utilizing a deterministic policy as the learning target when updating the critic network is sensitive to function approximation errors, resulting in elevated target estimate variance and imprecise value estimates \cite{lilicrap2016continuous}. To mitigate this issue during practical implementation, noise $\varepsilon$ is introduced by drawing it from a truncated normal distribution and adding it to the target actions:

\begin{equation}
\left\{\begin{array}{l}
\tilde{A} \longleftarrow \pi^{\prime}\left(s_{t+1} \mid \theta^{\pi^{\prime}}\right)+\varepsilon \\
\varepsilon \sim \operatorname{clip}(N(0, \sigma),-c, c), c>0
\end{array}\right.
\end{equation}

\noindent where $c$ is the truncation value, and $\sigma$ represents the standard deviation. The added noise is sampled from a normal distribution and then clipped to ensure proximity to the original action. In practical terms, strategies formulated with this methodology tend to be more conservative, assigning greater value to actions that exhibit resistance to disturbances.

The updating process involves both critic 1 and critic 2 networks by minimizing the TD-error, representing the gap between the estimated value and the target value. In the training phase, a data batch is sampled from the replay buffer. The critic network parameters are updated by minimizing the square of the TD-error:

\begin{equation} \label{Eq_update_Q_1}
\mathcal{L}_{c_i}\left(\theta^{Q_i}\right)=\mathbb{E}\left[y_j-Q_i\left(s_t, a_t \mid \theta^{Q_i}\right)\right]^2(i=1,2)
\end{equation}

\noindent where $y_j = r + \gamma \min_{i=1,2} Q_i'(s_{t+1}, a_{t+1}|\theta^{Q_i'})$ is target Q value, $s$ and $a$ denote the state and action, $r$ is reward.

The objective of the action behavior-determining policy network is to optimize the value of the value network, specifically improving control performance in the defined AD scenario addressed in this study. The actor network parameters can be updated utilizing the deterministic policy gradient method:

\begin{equation}\label{Eq_actor_loss}
\mathcal{L}_a\left(\theta^\pi\right)=\mathbb{E}\left(-Q\left(s_t, a_t \mid \theta^Q\right)\right)
\end{equation}

The update process for the target networks involves updating them using a soft update strategy. By introducing a learning rate $\tau$, a weighted average of the old target network parameters and the corresponding new network parameters is computed, and the resulting values are assigned to the target network:

\begin{equation}\label{Eq_update_target}
\left\{\begin{array}{l}
\theta^{Q^{\prime}} \leftarrow \tau \theta^Q+(1-\tau) \theta^{Q^{\prime}} \\
\theta^{\pi^{\prime}} \leftarrow \tau \theta^\pi+(1-\tau) \theta^{\pi^{\prime}}
\end{array}\right.
\end{equation}

To improve training stability, TD3 incorporates a delayed update mechanism, specifically involving the delayed update of the actor network—updating the actor network after multiple critic network updates \cite{fujimoto2018addressing}. This concept is intuitive as the actor network is updated based on maximizing the cumulative expected return and depends on the critic network for evaluation. When the critic network is highly unstable, the actor network is prone to oscillations.

% ---------------------------------------------------------------
\subsection{Reinforcement Learning with Human Guidance}

To integrate real-time guidance and feedback from humans into the RL algorithm, individuals have the authority to determine when to intervene, override the original actor actions and store them in the experience replay buffer \cite{10250993}. The human intervention guidance is characterized as a stochastic signal $I(s_t)$, which occurs based on the observation of the current state by the human driver. When human participants choose to take control of the RL training loop, they have full authority. Consequently, the agent's action $a_t$ can be represented as:

\begin{equation}\label{Eq_action}
a_t=I\left(s_t\right) \cdot a_t^{\text {H}}+\left[1-I\left(s_t\right)\right] \cdot a_t^{\mathrm{RL}}
\end{equation}

\noindent $a_t^{\mathrm{RL}}$ is the action given by the actor network. When there is no human guidance, $I(s_t)$ equals 0; when human intervention occurs, $I(s_t)$ equals 1.

Human guidance is also stored in the form of tuples in the experience replay buffer. The reconstructed experience replay pool facilitates the incorporation of human guidance into subsequent update processes. Upon occurrence of human guidance $a_t^{\text {H}}$, the TD3 algorithm's loss function needs to be adjusted to incorporate human experience. Consequently, the value network in the equation can be reformulated as:

\begin{equation}\label{Eq_update_Q}
\mathcal{L}_{c_i}\left(\theta^{Q_i}\right)=\mathbb{E}\left[y_j-Q_i\left(s_t, a_t^{\mathrm{H}} \mid \theta^{Q_i}\right)\right]^2(i=1,2)
\end{equation}

Certainly, as the value network updates depend on ${s_t, a_t^{\text{H}}}$ while the policy network continues to depend on ${\pi(s_t \mid \theta^{\pi})}$, it can result in inconsistent update directions for the actor and critic networks. Consequently, the loss function of the actor network in Eq. (\ref{Eq_actor_loss}) can be reformulated as:

\begin{equation}\label{Eq_update_pi}
\mathcal{L}_a\left(\theta^\pi\right)=\mathbb{E}\left[-Q\left(s_t, a_t \mid \theta^Q\right)+I\left(s_t\right) \cdot \omega_I \cdot\left(a_t-\pi\left(s_t \mid \theta^\pi\right)\right)\right]
\end{equation}

\noindent where $\omega_I$ serves as a constant factor that represents the weight of the uman-supervised loss. In Eq. (\ref{Eq_update_pi}), $a_t^{\mathrm{RL}}$ can be simply replaced with $a_t$, encompassing both human and RL policy actions. This aligns the updated direction with ${s_t, a_t^{\text{H}}}$ when human guidance occurs.

Next, we implement reward shaping based on human intervention, presuming that participants will intervene solely when the RL behavior proves unfavorable during training. In such cases, the intervention event can be considered a negative signal, and RL should avoid the corresponding state. This negative feedback can be achieved through reward shaping. According to the Eq. (\ref{Eq_action}), if a human performs an action, $I(s_t)$ equals 1, indicating that the agent generated a negative action at this time. Therefore, an additional penalty function can be used at time step $t$ to adjust the ordinary reward function:

\begin{equation}\label{Eq_reward_shape}
r_t^{\text {shape }}=r_t-r^{\text {penalty }} \cdot\left[\left(I\left(s_t\right)=1\right) \text { and }\left(I\left(s_{t-1}\right)=0\right)\right]
\end{equation}

\noindent where $r_t^{\text{shape}}$ is the reward after reshaping, $r^{\text {penalty}}$ is the penalty term at the moment of intervention, set as a constant.

Finally, we also adopted the priority replay mechanism from study \cite{9793564} in the experience replay buffer to prioritize learning human guidance data.

\begin{algorithm} 
    \caption{Fine tuning by RLHG}    
    \label{algorithm_1}       
    \begin{algorithmic}[1] 
    \Require Maximum episode number of RL $N_{ep}$, experience buffer $D$, the soft update coefficient $\tau$, the critic networks $Q_1$, $Q_2$ with parameter $\theta^{Q_1}$, $\theta^{Q_2}$; Load pre-trained transformer-enabled policy network $\pi$ with parameter $\theta^{\pi}$.  
    \Ensure Updated policy $\pi_{\theta}$.
    
    \For {episode = 1:$N_{ep}$}
        \State Observe the initial state $s_1$;
        \While {not done}
            \If {human intervened}
                \State Adopt human action $a_t = a^H_t$, set $I_t = 1$;
            \Else
                \State Execute $a_t$ = $\pi_{\theta}(s_t)$, set $I_t = 0$;
            \EndIf
            \State Get $s_{t+1}$, $r_t$ and store $(s_t,a_t,r_t,s_{t+1},I_t)$  to $D$;
            \State Shape reward using Eq. (\ref{Eq_reward_shape});
            \State Update critic networks using Eq. (\ref{Eq_update_Q_1}) or (\ref{Eq_update_Q});
            \If {delay end}
                \State Update actor network using Eq. (\ref{Eq_update_pi});
                \State Update target networks using Eq. (\ref{Eq_update_target});
            \EndIf
        \EndWhile
    \EndFor
   \end{algorithmic} 
\end{algorithm}

% === 4. Experiment =============================================================================
\textbf{\section{Experiment}}

\subsection{Driving Scenarios}

\begin{figure} [t!]
    \centering
    \subfloat[]{
        \includegraphics[width=0.49\linewidth]{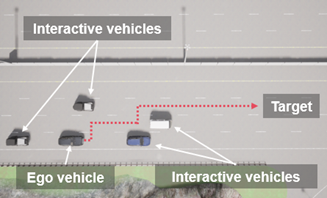}}
    \subfloat[]{
        \includegraphics[width=0.49\linewidth]{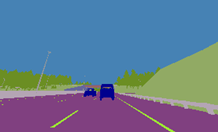}}\\
    \subfloat[]{
        \includegraphics[width=0.49\linewidth]{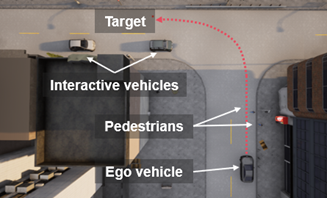}}
    \subfloat[]{
        \includegraphics[width=0.49\linewidth]{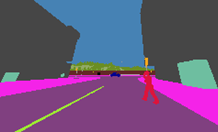}}
    \caption{The driving scenarios in CARLA: (a) Continuous overtaking scenario on highway; (b) Forward camera image of continuous overtaking scenario; (c) Unprotected left-turn scenario in city; (d) Forward camera image of left-turn scenario.}
    \label{fig2} 
\end{figure}

Compared to rule-based or model-based optimization methods, RL is more suitable for challenging driving tasks due to its excellent representational power and generalization ability. To evaluate the performance of RL, we selected two challenging scenarios (see Fig. \ref{fig2}).

% ==== FIG 2

\subsubsection{Continuous overtaking}
This scenario is depicted in Fig. \ref{fig2} (a) and (b). In this setup, the vehicle needs to complete a continuous overtaking task, spanning two lanes over a period. The AV first changes lanes from lane 1 to lane 2, then drives for a while before changing lanes again to lane 3. We assume that longitudinal control is handled by the intelligent driver model (IDM) [28], while lateral control is managed by the RL agent. In each episode, the types and initial velocities of surrounding vehicles vary within a certain range, controlled by the IDM to simulate following behavior.

\subsubsection{Unprotected left-turn}
Describes a scenario based on a busy city intersection with no traffic signals (Fig. \ref{fig2} (c) and (d)). In this scene, the AV needs to navigate through congested pedestrian traffic and then complete a left turn task without traffic signal protection. We assume that the lateral path of the AV is planned through other techniques, while vertical control is handled by the RL agent. In each episode, the initial positions and velocities of all traffic participants randomly vary within a certain range. The generated traffic flow is controlled by the IDM to simulate lane-keeping behavior. All surrounding drivers exhibit aggressive characteristics, indicating that they do not yield to the AV.

\subsection{RL-based Navigation Strategy}

In this section, RL algorithms from Section III are utilized to generate control modules by defining the state space, action space, and reward function.

\subsubsection{State variables}
Since the focus of this study is on the decision-making and planning performance of the perception backend, a single front-facing camera is used as input. Therefore, RGB channel front-view images are directly collected from the simulator and transformed into single-channel grayscale images, as shown in Fig. \ref{fig2} (b) and (d). At any time step $t$, time-aware is achieved by utilizing observations consisting of consecutive 2-frame images, represented as:

\begin{equation}
\mathbf{s}_t=\left[\chi_{t-1}, \chi_t\right]
\end{equation}

\noindent where $\chi \in \mathbb{R}^{80 \times 45}$ represents the image pixel matrix. In this setup, the agent, by considering historical state information, takes into account the consequences of past decisions when formulating the current decision. This configuration provides a solid theoretical foundation for achieving smoother control processes \cite{10164154}.

\subsubsection{Action variables}
The action variables are designed to facilitate AV in control tasks, where these variables can involve lateral or longitudinal commands tailored to specific needs. In the continuous overtaking scenario, the chosen action is the steering wheel angle for lateral control, while in the left-turn scenario, the selected action is the acceleration/brake pedal opening for longitudinal control. These actions are denoted as:

\begin{equation}
\mathbf{a}_t = \begin{cases}\alpha_t & \text {if overtake} \\ \delta_t & \text {if left-turn}\end{cases}
\end{equation}

\noindent where $\alpha$ is continuous steering commands, negative values indicate left turn commands, and positive values correspond to right turn commands. $\delta$ represents continuous pedal opening, with negative values indicating brake commands and positive values corresponding to acceleration commands.

\subsubsection{Reward function}

The objective of the AV is to navigate traffic scenarios efficiently while maintaining safe and smooth driving behavior. This is accomplished by formulating a driving strategy using RL and defining a suitable reward function. The objectives include navigating to the destination, avoiding collisions and deviations, maximizing driving efficiency, and maintaining smooth driving behavior. First, when the autonomous agent reaches the destination, it receives an immediate reward $r_{\text {success}}$:

\begin{equation}
r_{\text {success}}= \begin{cases}C_1 & \text { if success} \\ 0 & \text { otherwise }\end{cases}
\end{equation}

Secondly, when AV collides with other objects, self agency will receive a significant negative reward $r_{\text {collision}}$:

\begin{equation}
r_{\text {collision}}= \begin{cases}C_2 & \text { if collision} \\ 0 & \text { otherwise }\end{cases}
\end{equation}

Thirdly, when the AV deviates significantly from the driving route, it will receive a negative reward of $r_{\text {route-off}}$ as a punishment, as follows:

\begin{equation}
r_{\text {route-off}}= \begin{cases}C_3 & \text { if route-off} \\ 0 & \text { otherwise }\end{cases}
\end{equation}

Next, ego vehicle should arrive at the destination quickly. The reward $r_{\text {speed}}$ promoting higher speeds is defined as:

\begin{equation}
r_{\text {speed}}= \omega_1 \cdot v_{ego}
\end{equation}

Finally, larger instantaneous movements can bring discomfort to driving or riding, so we need to consider the comfort factor $r_{\text {comfort}}$ in the optimization goal, as shown below:

\begin{equation}
r_{\text {comfort }}= \begin{cases}\omega_2 \cdot|\alpha| & \text { if overtake } \\ \omega_3 \cdot\left|a c c_{\text {ego }}\right| & \text { if left-turn }\end{cases}
\end{equation}

The above reward functions aim to address multiple objectives simultaneously, including safety, efficiency, and comfort. The overall reward function is established by aggregating the aforementioned five sub-items:

\begin{equation}
r= r_{\text {success}} + r_{\text {collision}} + r_{\text {route-off}} + r_{\text {speed}} + r_{\text {comfort}}
\end{equation}

\subsection{Evaluation Metrics}

\subsubsection{Survival distance}
It calculates the distance that the AV can travel safely during each episode. A longer survival distance indicates that the system can better avoid potential hazards and obstacles, enhancing the safety of passengers and other road users.

\subsubsection{Collision rate}
This refers to the frequency of collisions during autonomous navigation. During testing, it computes the percentage of events where the vehicle collides with other vehicles, which is a crucial indicator of safety performance.

\subsubsection{Success rate}
This denotes the frequency at which the AV successfully completes the specified task or reaches the destination. It quantifies the percentage of events where the vehicle successfully reaches its goal. During testing, the success rate is measured over the total number of test runs.

\subsubsection{Intervention frequency}
This refers to the number of times manual intervention or correction is required during autonomous navigation. A lower intervention frequency indicates that the system can autonomously complete navigation tasks, reducing the burden on human drivers and improving the passenger experience. This study records the intervention frequency during training across different spans of episodes.

\subsection{Comparison Baselines}

To comprehensively evaluate the proposed method, our PTA-RLHG is compared with other existing DRL algorithms incorporating human factors. The baseline methods include:

\subsubsection{Hug-RL \cite{9793564}}
A state-of-the-art DRL algorithm involving human participation, utilizing mechanisms such as human supervision, intervention, and demonstration for assisted training. In scenarios without pre-training, this method has demonstrated superior learning efficiency and test performance compared to existing methods.

\subsubsection{RLfD}
This method combines demonstration data with RL, accelerating the RL process by observing expert-demonstrated behavior to enhance policy performance. It is based on the TD3 algorithm.

\subsubsection{Vanilla RL}
It adopts the traditional TD3 algorithm as a baseline, using the same representation for the state input.

\subsubsection{BC}
A deep neural network was constructed using the BC method to mimic the actions of human drivers through SL. This network shares the architecture with the policy network in Vanilla DRL.

\subsubsection{Transformer-enabled BC (TEBC)}
Trained similarly to the baseline (4), TEBC introduces a transformer module into the policy network. Through self-attention mechanisms, it compares and weights each token of the input sequence against others to obtain more comprehensive contextual information. This helps improve the model's understanding and modeling capabilities for different parts of the sequence.

For a fair comparison, the hyperparameters are set consistently, except for the learning rate of the strategy during the fine-tuning process, as detailed in Table \ref{tab_Parameters}.

% tab
\begin{table}[!htbp]
\caption{Hyperparameters for training process}
\label{tab_Parameters}
\begin{center}
\begin{tabular}{ccc}
 \hline
 Parameter & Description & Value \\ 
 \hline
 Maximum episode & Cutoff training episode number & 500 \\ 
 Replay buffer size & Capacity of the replay buffer & 38400\\ 
 Minibatch size & Capacity of minibatch & 128\\ 
 Learning rate decay & Delay of learning rate & 0.996\\  
 Soft updating factor & Update frequency to target networks & 1e-3\\ 
 Gamma ($\gamma$) & Discount factor & 0.95\\ 
 Actor learning rate & Fine-tuning process &  2e-6 \\ 
 Critic learning rate & Fine-tuning process &  5e-4 \\ 
 Actor learning rate & Reinforcement learning process &  2e-4 \\ 
 Critic learning rate & Reinforcement learning process &  5e-4 \\
 BC learning rate & Imitation learning process &  2e-4 \\ 
    
\hline
\end{tabular}
\end{center}
\end{table}

% ===  Results and Discussion =====================================================================
\textbf{\section{Results and Discussion}}

\subsection{Experimental Setting}

The proposed policy, along with various benchmarks, underwent training and evaluation on maps featuring both highway and regular urban roads within the CARLA simulator. To introduce real-world randomness, we randomized initial data, including the types of other traffic participants, their spawning locations, and initial velocities. Episodes concluded upon collision with objects or reaching the maximum duration. Initially, we assessed the training process of the algorithms in two scenarios, spanning 500 episodes. Notably, during the last 100 episodes, we considered the policy to have reached a mature level. To ensure an unbiased comparison, no human supervision occurred during this stage. Each policy underwent training with three random seeds for reliability. The trained policies were then evaluated in test scenarios, with the agent having blind spots in its perception area. The number of other traffic participants in test scenarios was set to 60-80\% of the training scenarios. We conducted 50 episodes of random scenario testing for policies trained with different random seeds, evaluating control performance using various metrics. The proposed policies and communication, control, and evaluation scripts were implemented in Python, with RL executed through the PyTorch toolkit. The simulation ran on a workstation featuring an Intel i7-13700KF CPU and Nvidia RTX4070Ti GPU.

\subsection{Training Results}

In this section, we assess the efficacy of our proposed method and juxtapose it with sophisticated baseline approaches. Moreover, we conduct ablation experiments to scrutinize the isolated effects of each module in our solution on overall performance.

% ==== FIG 3
\begin{figure} [t!]
\centering
   \subfloat[]{
       \includegraphics[width=0.49\linewidth]{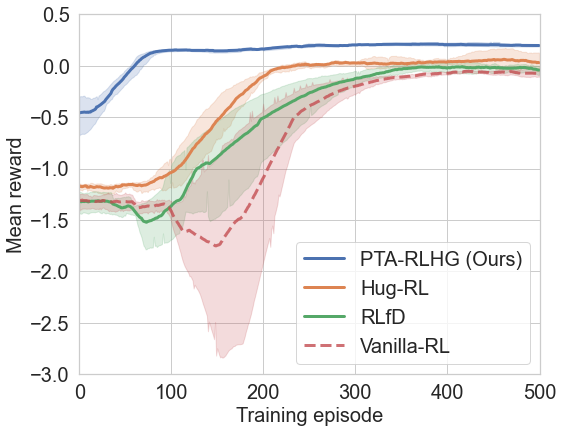}}
   \subfloat[]{
       \includegraphics[width=0.49\linewidth]{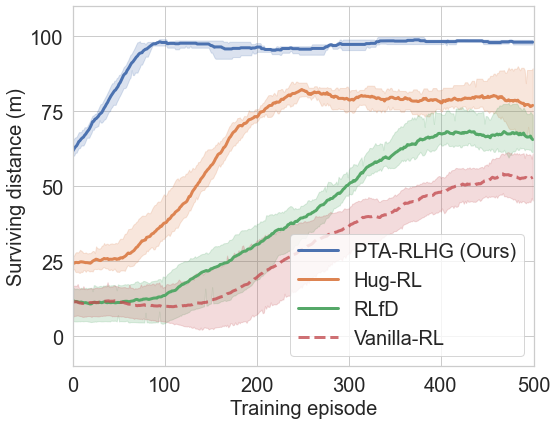}}\\
    \subfloat[]{
       \includegraphics[width=0.49\linewidth]{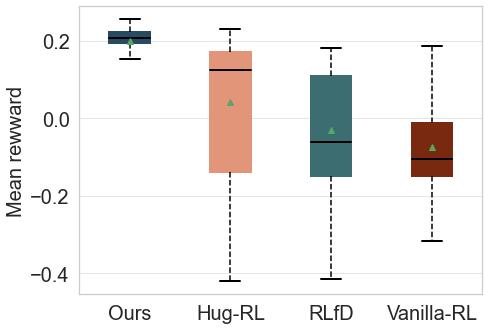}}
    \subfloat[]{
       \includegraphics[width=0.49\linewidth]{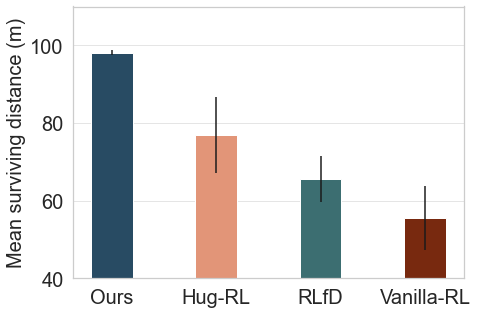}}
   \caption{The training process of our framework with baseline methods in overtake scenario: (a) Average reward; (b) Average driving distance; (c) Boxplot of rewards for the last 100 episodes; (d) Barplot of driving distances for the last 100 episodes.}
   \label{R_overtake} 
\end{figure}

Fig. \ref{R_overtake} (a) depicts the average reward curve during continuous overtaking scenario training, while Fig. \ref{R_overtake} (b) illustrates the distance traveled by the ego vehicle per episode. Both metrics, reflecting better learning outcomes with higher values, are presented with average and standard deviation error bars across three random seeds. Our proposed solution, compared to baseline methods, showcases remarkable training efficiency. The inclusion of pre-training notably boosts the agent's initial performance, emphasizing the positive impact of the policy initialization representation for rapid convergence. In baseline methods, Hug-RL demonstrates decent convergence efficiency but with limited asymptotic performance. Vanilla-RL experiences initial learning degradation, highlighting the effectiveness and necessity of human intervention and demonstration. Fig. \ref{R_overtake} (c) and (d) detail the distribution of average reward values and average travel distance over the final 100 training episodes, providing a fair comparison without human intervention. The reward distribution of PTA-RLHG is narrower and concentrated in higher ranges compared to other baselines, indicating the robustness of our method. Moreover, Fig. \ref{R_overtake} (c) shows that, in terms of mean asymptotic performance, PTA-RLHG improves by approximately 137.44\% compared to Vanilla-RL. The survival distance similarly indicates that our method matures gradually after RLHG fine-tuning, surpassing Hug-RL by about 21.48\% and Vanilla-RL by approximately 43.85\%.

% ==== FIG 4
\begin{figure} [t!]
\centering
   \subfloat[]{
       \includegraphics[width=0.49\linewidth]{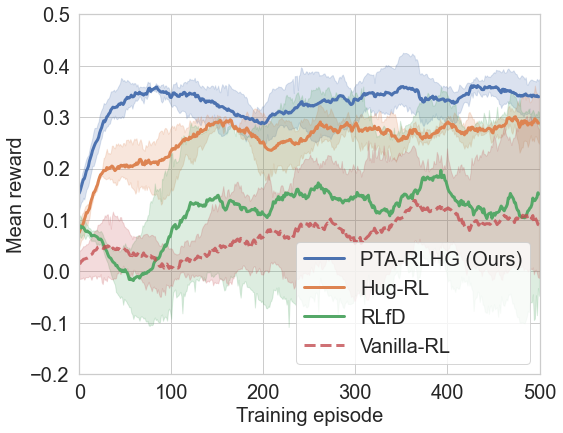}}
   \subfloat[]{
       \includegraphics[width=0.49\linewidth]{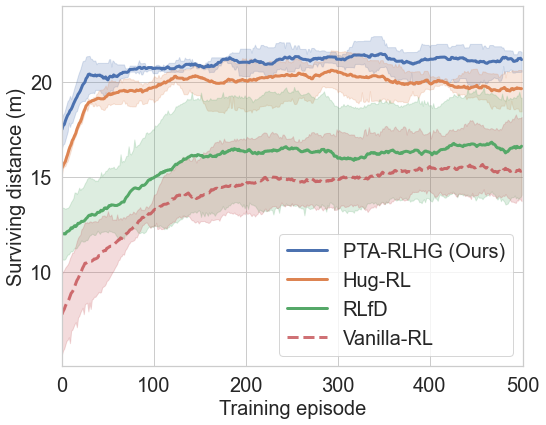}}\\
    \subfloat[]{
       \includegraphics[width=0.49\linewidth]{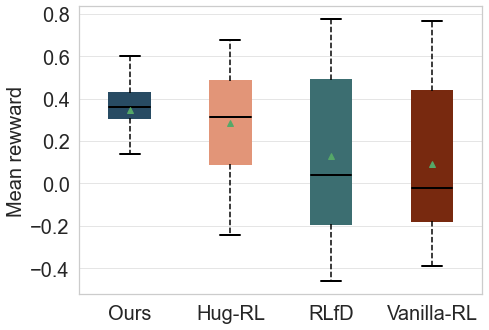}}
    \subfloat[]{
       \includegraphics[width=0.49\linewidth]{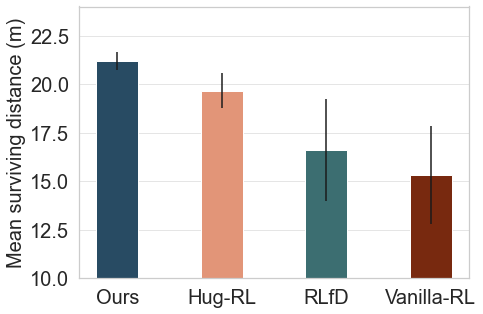}}
   \caption{The training process of our framework with baseline methods in left-turn scenario: (a) Average reward; (b) Average driving distance; (c) Boxplot of rewards for the last 100 episodes; (d) Barplot of driving distances for the last 100 episodes.}
   \label{R_leftturn} 
\end{figure}

Fig. \ref{R_leftturn} illustrates training scenarios involving an unprotected left-turn situation. Unlike lateral control, the intricate coordination between throttle and brake in longitudinal control presents a significant challenge to RL exploration. The complexity is further heightened by the introduction of pedestrians, adding to the difficulty of task completion in this scenario. From the visual representations, it is evident that PTA-RLHG consistently achieves the best asymptotic performance. The incorporation of human interaction enables the policy to swiftly learn through intervention and demonstration in the early stages of training. In contrast, the training process of RLfD indicates that relying solely on initial human demonstration data for enhancing policy training performance is insufficient. The presence of standard deviation error bars underscores the extreme instability in the training process when human intervention is absent.

% ==== FIG 5
\begin{figure} [t!]
\centering
   \subfloat[]{
       \includegraphics[width=0.49\linewidth]{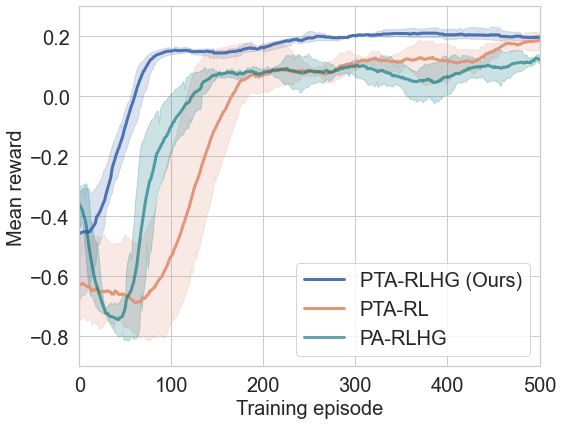}}
   \subfloat[]{
       \includegraphics[width=0.49\linewidth]{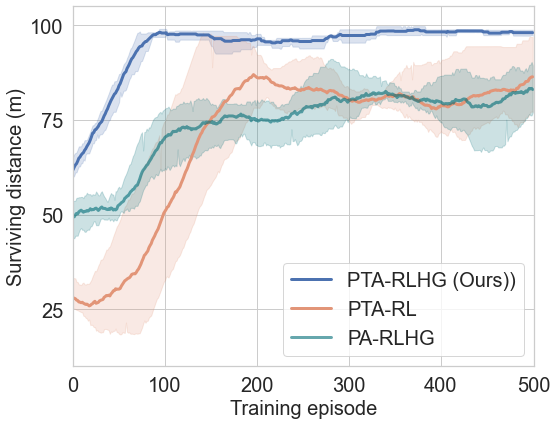}}\\
    \subfloat[]{
       \includegraphics[width=0.49\linewidth]{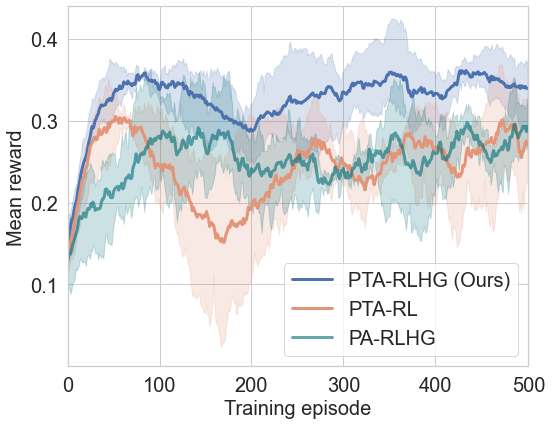}}
    \subfloat[]{
       \includegraphics[width=0.49\linewidth]{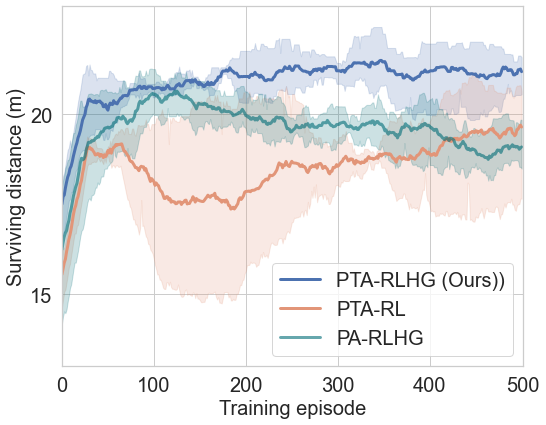}}
   \caption{The training process of our framework with ablation baseline methods in two scenarios: (a) Average reward for overtaking scenario; (b) Average driving distance for overtaking scenario; (c) Average reward for left-turn scenario; (d) Average driving distance for left-turn scenario.}
   \label{ablation} 
\end{figure}

To probe the influence of attention mechanisms and human guidance in our proposed method, we conducted ablation experiments, targeting two key components: 1) PTA-RL: Fine-tuning PTA using the conventional TD3 algorithm; 2) PA-RLHG: Removing the transformer module during pre-training, opting for a traditional convolutional encoder and MLP decoder structure, while maintaining RLHG during fine-tuning. All ablated baselines adhered to identical hyperparameters during fine-tuning. Fig. \ref{ablation} showcases the training outcomes in overtaking and left-turn scenarios. Despite a reduction in training efficiency and asymptotic performance, the ablated baseline methods exhibit overall convergence. Human intervention accelerates training efficiency in RLHG fine-tuning, while the absence of human guidance in PTA-RL may lead to a temporary decline in the early stages of training. This could be attributed to the vast and complex state space, prompting the agent to adopt random or suboptimal actions for exploration. While PA-RLHG demonstrates faster convergence efficiency initially, its asymptotic performance and later survival distance are constrained. This suggests that attention mechanisms play a pivotal role in enhancing policy stability by providing a better understanding of the scene.

% ==== FIG 6
\begin{figure} [t!]
\centering
   \subfloat[]{
       \includegraphics[width=0.9\linewidth]{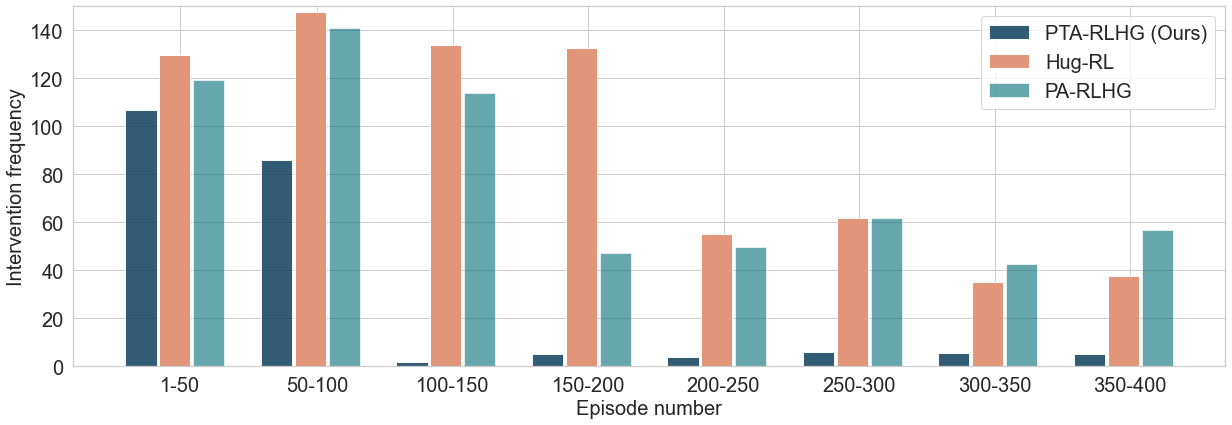}}\\
   \subfloat[]{
       \includegraphics[width=0.9\linewidth]{Fig/Fig7a.png}}
   \caption{The trend of intervention frequency during training process: (a) Continuous overtaking scenario; (b) Unprotected left-turn scenario.}
   \label{intervention} 
\end{figure}

% tab
\begin{table*}[!htbp]
\renewcommand\arraystretch{1.3}
\caption{Test results in two scenarios}
\label{Test}
\begin{center}
\begin{tabular}{cccccccc}
\hline
{}& \multicolumn{3}{c}{Overtake} &{}& \multicolumn{3}{c}{Left-turn} \\ 
\hline
Methods & Coll. (\%) & Succ. (\%) & Dis. (m) & {} & Coll. (\%) & Succ. (\%) & Dis. (m)\\ 

Vabilla-RL & 38.7 & 41.3 & 78.8 ± 12.56 &{}& 58.7 & 41.3 & 16.5 ± 2.32\\ 
 RLfD & 10.7 & 78.7 & 89.2 ± 6.02 &{}& 40.7 & 59.3 & 17.6 ± 2.49\\ 
 Hug-RL & 9.3 & 85.3 & 92.8 ± 2.98 &{}& 28.7 & 79.3 & 19.1 ± 1.05\\ 
 BC & 6.0 & 94.0 & 97.6 ± 0.04 &{}& 0 & 20.7 & 7.0 ± 1.81\\ 
 TEBC & 2.7 & 96.7 & 98.4 ± 1.96 &{}& 0 & 38.0 & 11.7 ± 1.84\\ 
  \hline
 PA-RLHG & 2.0 & 97.3 & 99.3 ± 0.32 &{}& 25.3 & 74.7 & 20.3 ± 0.60\\ 
 PTA-RL & 1.3 & 98.0 & 99.1 ± 0.90 &{}& 18.0 & 82.0 & 21.3 ± 0.93\\ 
\textbf{PTA-RLHG (Ours)}  & \textbf{0} & \textbf{100} & \textbf{99.8 ± 0.02} &{}& \textbf{4.0} & \textbf{94.0} & \textbf{22.8 ± 0.12}\\ 
\hline
\end{tabular}
\end{center}
\end{table*}

Introducing human guidance prevents the agent from encountering optimization challenges. To visually depict the learning progress of the expert strategy, we tracked the number of interventions within 400 episodes, as illustrated in Fig. \ref{intervention}. In the overtaking scenario, all methods show a gradual reduction in intervention, signifying the policies' progressive stabilization. Our method requires substantial human intervention only in the early stages of fine-tuning. Beyond 100 episodes, the policy rapidly stabilizes, necessitating minimal human intervention and exhibiting significantly lower intervention frequency than other baseline methods. In the left-turn scenario, our method still demonstrates a lower intervention frequency compared to the baselines. It's noteworthy that, in some cases, intervention frequencies exhibit a slight upward trend as training progresses, especially in the PA-RLHG method. This suggests that human intervention might introduce undesirable effects, possibly due to biases conflicting with action selection during the policy update process, leading to performance instability. However, our method highlights that incorporating attention mechanisms can mitigate this phenomenon to some extent.

\subsection{Testing Results}

To gauge the efficacy and robustness of well-trained policies, we conducted 50 tests across three different random seeds (with a total of 150 tests each strategy), comparing with both RL-based and IL methods (BC and TEBC) during testing. Table \ref{Test} summarizes key evaluation metrics: collision rate, success rate, and travel distance. Note that the sum of collision rate and success rate may not always equal 100\%, as the ego vehicle might occasionally reach an incorrect destination. The test results mirror the training outcomes, with the PTA-RLHG framework outperforming in both driving scenarios. Our method achieved the highest test success rate, markedly reducing collision failures. The two ablated baseline methods also demonstrated high success rates, underscoring the effectiveness of the pre-training and fine-tuning architecture. While IL-based methods performed well in overtaking scenarios, they faced challenges in left-turn longitudinal control tasks, displaying a collision rate of 0 alongside lower success rates and travel distances—a characteristic "freezing robot" dilemma. This could be attributed to BC-based methods relying solely on limited human demonstration data, lacking the exploratory capabilities needed for navigating unseen situations.

\begin{figure} [t!]
\centering
   \subfloat[]{
       \includegraphics[width=0.49\linewidth]{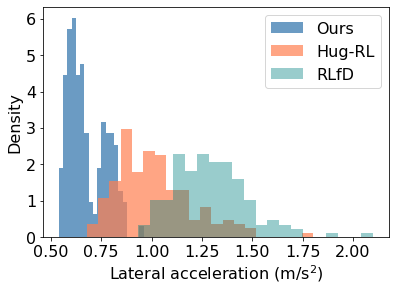}}
   \subfloat[]{
       \includegraphics[width=0.49\linewidth]{Fig/Fig8a.png}}
   \caption{Frequency distribution plot of the average absolute value
of the acceleration: (a) Lateral acceleration in the overtaking scenario; (b) Longitudinal acceleration in the left-turn scenario.}
   \label{acc} 
\end{figure}

These results collectively suggest that incorporating attention mechanisms in actor policy networks significantly enhances performance. Attention mechanisms enable focused attention on crucial information, facilitating more accurate decisions and action selections. The dynamic adjustment of attention allocation enhances adaptability to different environments and task requirements, boosting policy robustness and generalization capabilities. Through RL fine-tuning, pre-trained policies overcome distributional shift issues, surpassing IL performance limits and achieving superior results. Fig. \ref{acc} further illustrates the performance in terms of driving smoothness. Our proposed method exhibits the most favorable acceleration distribution, confirming its superiority in these aspects. Details about performance of our method are shown in \href{https://www.youtube.com/watch?v=grM0L9\_unI8}{video}.

% ==== FIG 7

% === ================= ===========================================================================
\textbf{\section{Conclusion}}

This study presents a novel end-to-end training framework for AV policies, utilizing Transformer and RLHG to improve sample efficiency and performance. The framework consists of two fundamental steps: firstly, integrating a Transformer module into the policy network to identify pertinent subspaces or features correlated with the current state. This is complemented by pre-training via BC. Secondly, fine-tuning is executed using RLHG, harnessing human guide to make the model in acquiring more precise and safe driving decisions.

We extensively validated our proposed method in two challenging simulated driving scenarios, showcasing superior performance in sample efficiency, robustness, safety, and driving smoothness through both quantitative and qualitative analyses. Our investigation into framework components highlighted the transformative impact of the Transformer's importance-capturing mechanism, crucial for policy stabilization and enhancement, enabling flexible adaptation to diverse environments and tasks. Additionally, fine-tuning with RLHG significantly enhanced agent adaptability in the two challenging driving tasks.

With this innovative training approach, we anticipate achieving significant breakthroughs in the field of AD, realizing higher levels of driving performance and intelligence. Despite these advancements, we acknowledge limitations, such as the use of a single forward-facing camera causing perceptual blind spots. Addressing the burden of online interventions during fine-tuning, exploring the complexity of pre-trained networks and multimodal fusion are essential areas for future studies. Additionally, considering training methods for generative pre-trained Transformer in research plans holds potential. Results from these ongoing studies will be reported in due course.

\textbf{
\bibliographystyle{IEEEtran}}
\small\bibliography{Bibliography}

\begin{thebibliography}{10}
\providecommand{\url}[1]{#1}
\csname url@rmstyle\endcsname
\providecommand{\newblock}{\relax}
\providecommand{\bibinfo}[2]{#2}
\providecommand\BIBentrySTDinterwordspacing{\spaceskip=0pt\relax}
\providecommand\BIBentryALTinterwordstretchfactor{4}
\providecommand\BIBentryALTinterwordspacing{\spaceskip=\fontdimen2\font plus
\BIBentryALTinterwordstretchfactor\fontdimen3\font minus \fontdimen4\font\relax}
\providecommand\BIBforeignlanguage[2]{{%
\expandafter\ifx\csname l@#1\endcsname\relax
\typeout{** WARNING: IEEEtran.bst: No hyphenation pattern has been}%
\typeout{** loaded for the language `#1'. Using the pattern for}%
\typeout{** the default language instead.}%
\else
\language=\csname l@#1\endcsname
\fi
#2}}

\bibitem{chib2023recent}
P.~S. Chib and P.~Singh, ``Recent advancements in end-to-end autonomous driving using deep learning: A survey,'' \emph{IEEE Transactions on Intelligent Vehicles}, 2023.

\bibitem{feng2023dense}
S.~Feng, H.~Sun, X.~Yan, H.~Zhu, Z.~Zou, S.~Shen, and H.~X. Liu, ``Dense reinforcement learning for safety validation of autonomous vehicles,'' \emph{Nature}, vol. 615, no. 7953, pp. 620--627, 2023.

\bibitem{codevilla2018end}
F.~Codevilla, M.~M{\"u}ller, A.~L{\'o}pez, V.~Koltun, and A.~Dosovitskiy, ``End-to-end driving via conditional imitation learning,'' in \emph{2018 IEEE international conference on robotics and automation (ICRA)}.\hskip 1em plus 0.5em minus 0.4em\relax IEEE, 2018, pp. 4693--4700.

\bibitem{codevilla2019exploring}
F.~Codevilla, E.~Santana, A.~M. L{\'o}pez, and A.~Gaidon, ``Exploring the limitations of behavior cloning for autonomous driving,'' in \emph{Proceedings of the IEEE/CVF International Conference on Computer Vision}, 2019, pp. 9329--9338.

\bibitem{zou2021deep}
Q.~Zou, K.~Xiong, Q.~Fang, and B.~Jiang, ``Deep imitation reinforcement learning for self-driving by vision,'' \emph{CAAI Transactions on Intelligence Technology}, vol.~6, no.~4, pp. 493--503, 2021.

\bibitem{shi2023efficient}
J.~Shi, T.~Zhang, J.~Zhan, S.~Chen, J.~Xin, and N.~Zheng, ``Efficient lane-changing behavior planning via reinforcement learning with imitation learning initialization,'' in \emph{2023 IEEE Intelligent Vehicles Symposium (IV)}.\hskip 1em plus 0.5em minus 0.4em\relax IEEE, 2023, pp. 1--8.

\bibitem{wu2023brief}
T.~Wu, S.~He, J.~Liu, S.~Sun, K.~Liu, Q.-L. Han, and Y.~Tang, ``A brief overview of chatgpt: The history, status quo and potential future development,'' \emph{IEEE/CAA Journal of Automatica Sinica}, vol.~10, no.~5, pp. 1122--1136, 2023.

\bibitem{wang2023bevgpt}
P.~Wang, M.~Zhu, H.~Lu, H.~Zhong, X.~Chen, S.~Shen, X.~Wang, and Y.~Wang, ``Bevgpt: Generative pre-trained large model for autonomous driving prediction, decision-making, and planning,'' \emph{arXiv preprint arXiv:2310.10357}, 2023.

\bibitem{xu2023drivegpt4}
Z.~Xu, Y.~Zhang, E.~Xie, Z.~Zhao, Y.~Guo, K.~K. Wong, Z.~Li, and H.~Zhao, ``Drivegpt4: Interpretable end-to-end autonomous driving via large language model,'' \emph{arXiv preprint arXiv:2310.01412}, 2023.

\bibitem{vaswani2017attention}
A.~Vaswani, N.~Shazeer, N.~Parmar, J.~Uszkoreit, L.~Jones, A.~N. Gomez, {\L}.~Kaiser, and I.~Polosukhin, ``Attention is all you need,'' \emph{Advances in neural information processing systems}, vol.~30, 2017.

\bibitem{dosovitskiy2020image}
A.~Dosovitskiy, L.~Beyer, A.~Kolesnikov, D.~Weissenborn, X.~Zhai, T.~Unterthiner, M.~Dehghani, M.~Minderer, G.~Heigold, S.~Gelly, \emph{et~al.}, ``An image is worth 16x16 words: Transformers for image recognition at scale,'' \emph{arXiv preprint arXiv:2010.11929}, 2020.

\bibitem{qian2021blending}
S.~Qian, H.~Shao, Y.~Zhu, M.~Li, and J.~Jia, ``Blending anti-aliasing into vision transformer,'' \emph{Advances in Neural Information Processing Systems}, vol.~34, pp. 5416--5429, 2021.

\bibitem{gao2020vectornet}
J.~Gao, C.~Sun, H.~Zhao, Y.~Shen, D.~Anguelov, C.~Li, and C.~Schmid, ``Vectornet: Encoding hd maps and agent dynamics from vectorized representation,'' in \emph{Proceedings of the IEEE/CVF Conference on Computer Vision and Pattern Recognition}, 2020, pp. 11\,525--11\,533.

\bibitem{huang2023differentiable}
Z.~Huang, H.~Liu, J.~Wu, and C.~Lv, ``Differentiable integrated motion prediction and planning with learnable cost function for autonomous driving,'' \emph{IEEE transactions on neural networks and learning systems}, 2023.

\bibitem{huang2023conditional}
------, ``Conditional predictive behavior planning with inverse reinforcement learning for human-like autonomous driving,'' \emph{IEEE Transactions on Intelligent Transportation Systems}, 2023.

\bibitem{gou2022driver}
C.~Gou, Y.~Zhou, and D.~Li, ``Driver attention prediction based on convolution and transformers,'' \emph{The Journal of Supercomputing}, vol.~78, no.~6, pp. 8268--8284, 2022.

\bibitem{meinhardt2022trackformer}
T.~Meinhardt, A.~Kirillov, L.~Leal-Taixe, and C.~Feichtenhofer, ``Trackformer: Multi-object tracking with transformers,'' in \emph{Proceedings of the IEEE/CVF conference on computer vision and pattern recognition}, 2022, pp. 8844--8854.

\bibitem{chitta2022transfuser}
K.~Chitta, A.~Prakash, B.~Jaeger, Z.~Yu, K.~Renz, and A.~Geiger, ``Transfuser: Imitation with transformer-based sensor fusion for autonomous driving,'' \emph{IEEE Transactions on Pattern Analysis and Machine Intelligence}, 2022.

\bibitem{chitta2021neat}
K.~Chitta, A.~Prakash, and A.~Geiger, ``Neat: Neural attention fields for end-to-end autonomous driving,'' in \emph{Proceedings of the IEEE/CVF International Conference on Computer Vision}, 2021, pp. 15\,793--15\,803.

\bibitem{shao2023safety}
H.~Shao, L.~Wang, R.~Chen, H.~Li, and Y.~Liu, ``Safety-enhanced autonomous driving using interpretable sensor fusion transformer,'' in \emph{Conference on Robot Learning}.\hskip 1em plus 0.5em minus 0.4em\relax PMLR, 2023, pp. 726--737.

\bibitem{shao2023reasonnet}
H.~Shao, L.~Wang, R.~Chen, S.~L. Waslander, H.~Li, and Y.~Liu, ``Reasonnet: End-to-end driving with temporal and global reasoning,'' in \emph{Proceedings of the IEEE/CVF Conference on Computer Vision and Pattern Recognition}, 2023, pp. 13\,723--13\,733.

\bibitem{wen2023dilu}
L.~Wen, D.~Fu, X.~Li, X.~Cai, T.~Ma, P.~Cai, M.~Dou, B.~Shi, L.~He, and Y.~Qiao, ``Dilu: A knowledge-driven approach to autonomous driving with large language models,'' \emph{arXiv preprint arXiv:2309.16292}, 2023.

\bibitem{wang2023drivedreamer}
X.~Wang, Z.~Zhu, G.~Huang, X.~Chen, and J.~Lu, ``Drivedreamer: Towards real-world-driven world models for autonomous driving,'' \emph{arXiv preprint arXiv:2309.09777}, 2023.

\bibitem{mnih2015human}
V.~Mnih, K.~Kavukcuoglu, D.~Silver, A.~A. Rusu, J.~Veness, M.~G. Bellemare, A.~Graves, M.~Riedmiller, A.~K. Fidjeland, G.~Ostrovski, \emph{et~al.}, ``Human-level control through deep reinforcement learning,'' \emph{nature}, vol. 518, no. 7540, pp. 529--533, 2015.

\bibitem{lilicrap2016continuous}
T.~Lilicrap, J.~Hunt, A.~Pritzel, N.~Hess, T.~Erez, D.~Silver, Y.~Tassa, and D.~Wiestra, ``Continuous control with deep reinforcement learning,'' in \emph{International Conference on Representation Learning (ICRL)}, 2016.

\bibitem{HU2024130097}
D.~Hu, C.~Huang, G.~Yin, Y.~Li, Y.~Huang, H.~Huang, J.~Wu, W.~Li, and H.~Xie, ``A transfer-based reinforcement learning collaborative energy management strategy for extended-range electric buses with cabin temperature comfort consideration,'' \emph{Energy}, vol. 290, p. 130097, 2024.

\bibitem{hu2023asynchronous}
Z.~Hu, X.~Gao, K.~Wan, Q.~Wang, and Y.~Zhai, ``Asynchronous curriculum experience replay: A deep reinforcement learning approach for uav autonomous motion control in unknown dynamic environments,'' \emph{IEEE Transactions on Vehicular Technology}, 2023.

\bibitem{wu2022uncertainty}
J.~Wu, Z.~Huang, and C.~Lv, ``Uncertainty-aware model-based reinforcement learning: Methodology and application in autonomous driving,'' \emph{IEEE Transactions on Intelligent Vehicles}, vol.~8, no.~1, pp. 194--203, 2022.

\bibitem{HU2023121227}
D.~Hu, H.~Xie, K.~Song, Y.~Zhang, and L.~Yan, ``An apprenticeship-reinforcement learning scheme based on expert demonstrations for energy management strategy of hybrid electric vehicles,'' \emph{Applied Energy}, vol. 342, p. 121227, 2023.

\bibitem{liu2022improved}
H.~Liu, Z.~Huang, J.~Wu, and C.~Lv, ``Improved deep reinforcement learning with expert demonstrations for urban autonomous driving,'' in \emph{2022 IEEE Intelligent Vehicles Symposium (IV)}.\hskip 1em plus 0.5em minus 0.4em\relax IEEE, 2022, pp. 921--928.

\bibitem{pfeiffer2018reinforced}
M.~Pfeiffer, S.~Shukla, M.~Turchetta, C.~Cadena, A.~Krause, R.~Siegwart, and J.~Nieto, ``Reinforced imitation: Sample efficient deep reinforcement learning for mapless navigation by leveraging prior demonstrations,'' \emph{IEEE Robotics and Automation Letters}, vol.~3, no.~4, pp. 4423--4430, 2018.

\bibitem{tai2018socially}
L.~Tai, J.~Zhang, M.~Liu, and W.~Burgard, ``Socially compliant navigation through raw depth inputs with generative adversarial imitation learning,'' in \emph{2018 IEEE international conference on robotics and automation (ICRA)}.\hskip 1em plus 0.5em minus 0.4em\relax IEEE, 2018, pp. 1111--1117.

\bibitem{huang2023goal}
W.~Huang, Y.~Zhou, X.~He, and C.~Lv, ``Goal-guided transformer-enabled reinforcement learning for efficient autonomous navigation,'' \emph{arXiv preprint arXiv:2301.00362}, 2023.

\bibitem{wu2021deep}
Y.~Wu, S.~Liao, X.~Liu, Z.~Li, and R.~Lu, ``Deep reinforcement learning on autonomous driving policy with auxiliary critic network,'' \emph{IEEE transactions on neural networks and learning systems}, 2021.

\bibitem{WU202375}
J.~Wu, Z.~Huang, Z.~Hu, and C.~Lv, ``Toward human-in-the-loop ai: Enhancing deep reinforcement learning via real-time human guidance for autonomous driving,'' \emph{Engineering}, vol.~21, pp. 75--91, 2023.

\bibitem{9793564}
J.~Wu, Z.~Huang, W.~Huang, and C.~Lv, ``Prioritized experience-based reinforcement learning with human guidance for autonomous driving,'' \emph{IEEE Transactions on Neural Networks and Learning Systems}, vol.~35, no.~1, pp. 855--869, 2024.

\bibitem{griffith2013policy}
S.~Griffith, K.~Subramanian, J.~Scholz, C.~L. Isbell, and A.~L. Thomaz, ``Policy shaping: Integrating human feedback with reinforcement learning,'' \emph{Advances in neural information processing systems}, vol.~26, 2013.

\bibitem{knox2011augmenting}
W.~B. Knox and P.~Stone, ``Augmenting reinforcement learning with human feedback,'' in \emph{ICML 2011 Workshop on New Developments in Imitation Learning (July 2011)}, vol. 855, 2011, p.~3.

\bibitem{9578103}
A.~Prakash, K.~Chitta, and A.~Geiger, ``Multi-modal fusion transformer for end-to-end autonomous driving,'' in \emph{2021 IEEE/CVF Conference on Computer Vision and Pattern Recognition (CVPR)}, 2021, pp. 7073--7083.

\bibitem{fujimoto2018addressing}
S.~Fujimoto, H.~Hoof, and D.~Meger, ``Addressing function approximation error in actor-critic methods,'' in \emph{International conference on machine learning}.\hskip 1em plus 0.5em minus 0.4em\relax PMLR, 2018, pp. 1587--1596.

\bibitem{10250993}
J.~Wu, Y.~Zhou, H.~Yang, Z.~Huang, and C.~Lv, ``Human-guided reinforcement learning with sim-to-real transfer for autonomous navigation,'' \emph{IEEE Transactions on Pattern Analysis and Machine Intelligence}, vol.~45, no.~12, pp. 14\,745--14\,759, 2023.

\bibitem{10164154}
J.~Wu, Z.~Song, and C.~Lv, ``Deep reinforcement learning based energy-efficient decision-making for autonomous electric vehicle in dynamic traffic environments,'' \emph{IEEE Transactions on Transportation Electrification}, pp. 1--1, 2023.

\end{thebibliography}

\end{document}